\documentclass[]{interact}

\usepackage{color}
\usepackage[dvipsnames]{xcolor}

\usepackage{epstopdf}
\usepackage[caption=false]{subfig}

\usepackage[numbers,sort&compress]{natbib}
\bibpunct[, ]{[}{]}{,}{n}{,}{,}
\renewcommand\bibfont{\fontsize{10}{12}\selectfont}
\makeatletter
\def\NAT@def@citea{\def\@citea{\NAT@separator}}
\makeatother

\theoremstyle{plain}

\theoremstyle{definition}

\theoremstyle{remark}

\usepackage{multirow}
\usepackage{graphicx}
\usepackage{tabularx}

\usepackage{url} 

\begin{document}

\title{Co-Scale Cross-Attentional Transformer for \\Rearrangement Target Detection}

\author{
\name{Haruka Matsuo\thanks{CONTACT Haruka Matsuo. Email: haruka.matsuo-25@keio.jp}\thanks{This article has been accepted for publication in Advanced Robotics, published by Taylor \& Francis.}, Shintaro Ishikawa and Komei Sugiura}
\affil{Keio University, 3-14-1 Hiyoshi, Kohoku, Yokohama, Kanagawa 223-8522, Japan}
}

\maketitle

\begin{abstract}
Rearranging objects (e.g. vase, door) back in their original positions is one of the most fundamental skills for domestic service robots (DSRs).
In rearrangement tasks, it is crucial to detect the objects that need to be rearranged according to the goal and current states.
In this study, we focus on Rearrangement Target Detection (RTD), where the model generates a change mask for objects that should be rearranged.
Although many studies have been conducted in the field of Scene Change Detection (SCD), most SCD methods often fail to segment objects with complex shapes and fail to detect the change in the angle of objects that can be opened or closed.
In this study, we propose a Co-Scale Cross-Attentional Transformer for RTD.
We introduce the Serial Encoder which consists of a sequence of serial blocks and the Cross-Attentional Encoder which models the relationship between the goal and current states.
We built a new dataset consisting of RGB images and change masks regarding the goal and current states.
We validated our method on the dataset and the results demonstrated that our method outperformed baseline methods on $F_1$-score and mean IoU.
\end{abstract}


\vspace{-0.8mm}
\section{Introduction
\label{intro}
}
\vspace{-0.7mm}

In our aging society, the shortage of home caregivers has become a social problem, and domestic service robots (DSRs) are expected to solve this problem \cite{yamamoto2019development}.
It would be useful if a DSR could handle a rearrangement task, where it recovers a goal room configuration after its user has moved or changed the state of several objects from a current room configuration \cite{batra2020rearrangement}.
In the rearrangement tasks, it is crucial to detect the objects that need to be rearranged based on the comparison between the original goal and current states.
If such objects can be detected, the objects can be successfully rearranged in the rearrangement task.
However, the detection performance is currently insufficient.

In this study, we focus on RTD, where a model detects objects that need to be rearranged based on the goal and current states. 
For example, if an object located at the edge of the desk in the goal state has been moved to the center of the desk in the current state, the model should detect the object and output an image where the object is masked.
Moreover, if a door that was closed in the goal state is open in the current state, the model should output an image where the door is masked.

RTD is challenging because simple approaches based on pixel-level differences between the two input images often fail to segment the objects that should be rearranged.
Using such an approach for RTD can result in segmentation errors, especially when the robot pose in the current state is not identical to that in the goal state.
These errors are caused by a few significant differences at the pixel level or changes in light and shadows.

\begin{figure}[t]
    \centering
        \begin{minipage}{\hsize}
            \centering
            \includegraphics[scale=0.325]{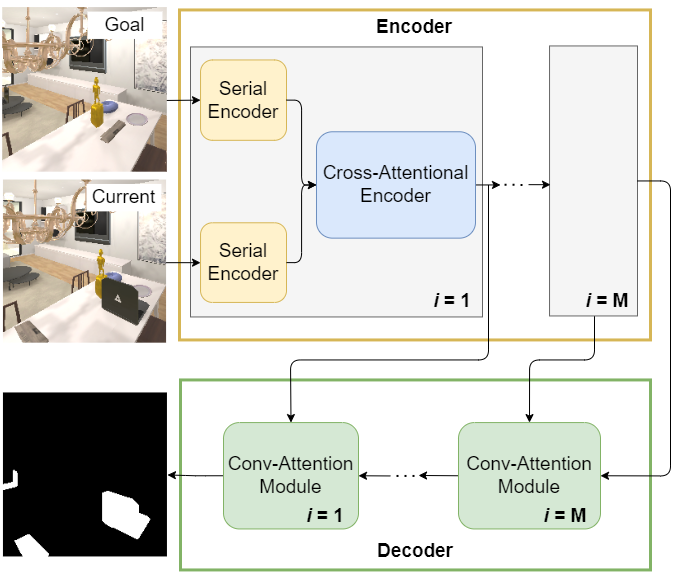}
        \end{minipage}
    \vspace{-2mm}
    \caption{\normalsize Our method overview: Given images of the goal and current states, our method generates a change mask for objects that should be rearranged.}
    \label{fig:eye-catch}
\end{figure}

There have been many studies on SCD \cite{sakurada2020weakly, chen2021dr}, which is closely related to RTD.
However, most of them often fail to segment objects with complex shapes or detect changes in the angle of doors.

In this paper, we propose a Co-Scale Cross-Attentional Transformer\footnote{\url{https://github.com/keio-smilab24/Co-Scale_Cross-Attentional_Transformer}} that detects the rearrangement of target objects based on images of the goal and current states.
We introduce the Serial Encoder that consists of a sequence of CoaT serial blocks to enhance visual information.
We also introduce the Cross-Attentional Encoder that models the relationship between the goal and current states using cross-attention to generate an appropriate change mask.
The difference compared with existing methods is that our method introduces the Serial Encoder using a CoaT serial block \cite{xu2021co} for two lanes in parallel instead of one lane and the Cross-Attentional Encoder that models the relationship between the goal and current states.
Our method is expected to correctly segment objects with complex shapes and the change in the angles of doors by using the attention mechanisms.

The main contributions of this paper are summarized as follows:
\begin{itemize}
    \item We propose the Co-Scale Cross-Attentional Transformer for RTD.
    \item We introduce the Serial Encoder that extracts features using a CoaT serial block \cite{xu2021co} for two lanes in parallel, the goal and current states, instead of one lane.
    \item We introduce the Cross-Attentional Encoder that models the relationship between image features in the goal and current states.
\end{itemize}

\vspace{2mm}
\section{Related Work
\label{related}
}

\vspace{-0.5mm}
\subsection{Change Detection}
\vspace{-0.5mm}

Numerous studies have been conducted in the field of change detection \cite{Varghese2018changenet, jiang2020pga, shi2022dsamnet, park2022dual}.
\cite{shi2020change} is a survey paper in the field of change detection.
It provides a comprehensive survey of the methods, standard datasets and issues in change detection, and classifies the methods based on their application domains.

The field of change detection can be divided into two types: Remote Sensing Change Detection (RS) and Scene Change Detection (SCD).
RS is aimed at detecting changes in buildings, roads, and farmland based on satellite and aerial images (e.g., \cite{lyu2018long, peng2020semicdnet, bandara2022changef}).
ChangeFormer \cite{bandara2022changef} is a model that unifies a hierarchically structured transformer encoder with an MLP decoder in a Siamese network architecture.
In \cite{yan2022fully} and \cite{chen2021remote}, Transformer-based change segmentation methods are proposed. Fully Transformer Network \cite{yan2022fully}  incorporates Swin Transformer \cite{liu2021swin} as part of its methodology without modifying the Fully Transformer Network architecture, and similarly, the bitemporal image transformer \cite{chen2021remote} does not adopt the Transformer architecture but integrates Transformer Encoder and Decoder separately within its approach.

SCD is aimed at detecting changes in streetscapes or objects based on street view images or indoor images (e.g., \cite{sakurada2020weakly, chen2021dr, park2022dual}).
CSCDNet \cite{sakurada2020weakly} is a model that has a novel Siamese network structure and introduces a correlation layer to deal with the difference in camera viewpoints.

The standard datasets used in change detection include PCD \cite{sakurada2015pcd}, VL-CMU-CD \cite{alcantarilla2018street}, and ChangeSim \cite{park2021changesim}.
VL-CMU-CD \cite{alcantarilla2018street} is a dataset built in an urban environment with seasonal and lighting variations. It consists of 1,362 image pairs obtained from 152 street view sequences. Each sequence contains approximately 9 frames.
PCD \cite{sakurada2015pcd} consists of two subsets, ``TSUNAMI'' and ``GSV.'' Each subset consists of 100 panoramic image pairs.
ChangeSim \cite{park2021changesim} is a dataset collected from industrial indoor simulation environments. It consists of approximately 130,000 image pairs obtained from 80 sequences. Each sequence contains 500 to 6,500 frames in 10 different environments.

Unlike most SCD methods, such as CSCDNet \cite{sakurada2020weakly}, our method can detect changes in the angle of doors or drawers.
Moreover, the change detection performance of CSCDNet \cite{sakurada2020weakly}, which concatenates image features extracted by ResNet \cite{he2016deep} is insufficient to model complex relationships between the goal and current states.
On the contrary, our method introduces the Serial Encoder that uses CoaT serial blocks \cite{xu2021co} for two lanes in parallel instead of one lane and the Cross-Attentional Encoder that uses cross-attention.

\vspace{-0.5mm}
\subsection{Rearrangement}
\vspace{-0.5mm}

Many studies have also been conducted in the field of rearrangement tasks \cite{krontiris2014rearranging, king2017unobservable, shome2021synchronized, RoomR, trabucco2023a}.
\cite{batra2020rearrangement} provides a comprehensive overview of the rearrangement tasks, and it describes conventional methods, a general framework, rearrangement scenarios, and metrics.
\cite{king2017unobservable} proposes a model that generates open-loop trajectories to solve rearrangement planning problems using nonprehensile manipulation.
\cite{trabucco2023a} proposes a model that uses voxel-based semantic maps to infer what should be rearranged compared with the goal state.

In addition to RTD, there are tasks that DSRs perform based on images obtained in the home environment \cite{matsubara22corsmal, magassouba2021predicting, iocchi2015robocup}.
\cite{matsubara22corsmal} proposes a mask-based geometric algorithm for best frame selection to estimate 3D models of containers.
PonNet \cite{magassouba2021predicting} is a model that visualizes the risk of damaging collisions based only on visual inputs.
RoboCup@Home competitions have been conducted to benchmark physical robots that tidy up objects, where each goal state is not given as an image \cite{iocchi2015robocup}.



\section{Problem Statement
\label{sec:problem}
}

\begin{figure}[t]
    \vspace{2.7mm}
    \centering
    \includegraphics[width=0.95\linewidth]{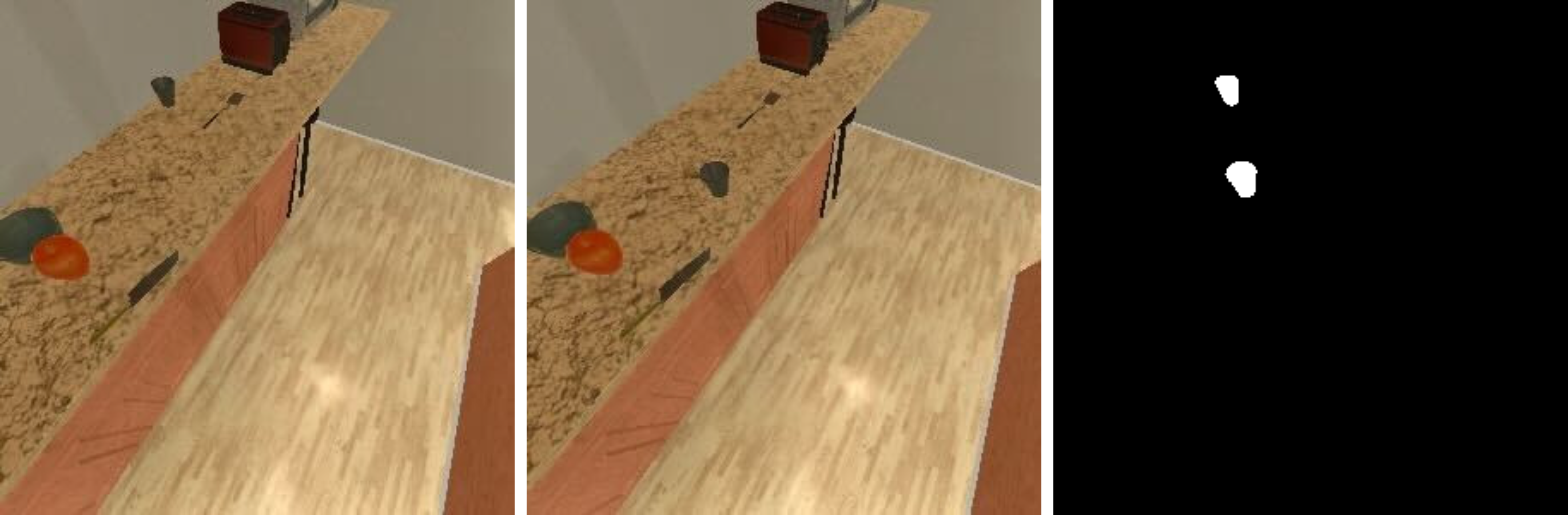}
    \vspace{-3mm}
    \caption{\normalsize An example of RTD. From left to right: goal state, current state, and change mask.}
    \label{fig:problem_example}
\end{figure}

In this paper, we focus on RTD, in which the model detects rearranged objects between the goal and current states of various indoor environments.
In the task, the robot identifies randomly moved objects based on prior observation of the goal state.
Fig.~\ref{fig:problem_example} shows an example of RTD.
The left, middle, and right images show the goal state, current state, and change mask, respectively.
Given the images of the goal and current states, the model should detect objects whose positions have changed.
The output of the model should be an image masking the objects.

The task is characterized as follows:
\begin{itemize}
    \item \textbf{Input:} RGB images of the goal and current states.
    \item \textbf{Output:} A change mask.
\end{itemize}

The terminology used in this paper is defined as follows:
\begin{itemize}
    \item \textbf{Rearrangement target object:} Object whose position and orientation have been changed. Drawers and doors that have changed their angles, between the goal and current states.
\end{itemize}

Since our goal is to detect rearranged objects, we do not handle category recognition and rearrangement of objects.
We also assume that the number of rearranged objects in an image is less than five, following the settings of the 2022 AI2-THOR Rearrangement Challenge \cite{RoomR}.
We use the mean Intersection over Union (mIoU) and $F_1$-score as the evaluation metrics.
We describe those metrics in detail in Section \textcolor{blue}{5}.
Training a model often requires a large amount of training data.
However, data collection with physical robots is labor-intensive because it requires a human to place objects. 
Therefore, we adopt a simulation environment to efficiently collect the data.

\section{Proposed Method
\label{method}
}

\begin{figure*}[t]
    \vspace{2.8mm}
    \centering
        \includegraphics[width=\linewidth]{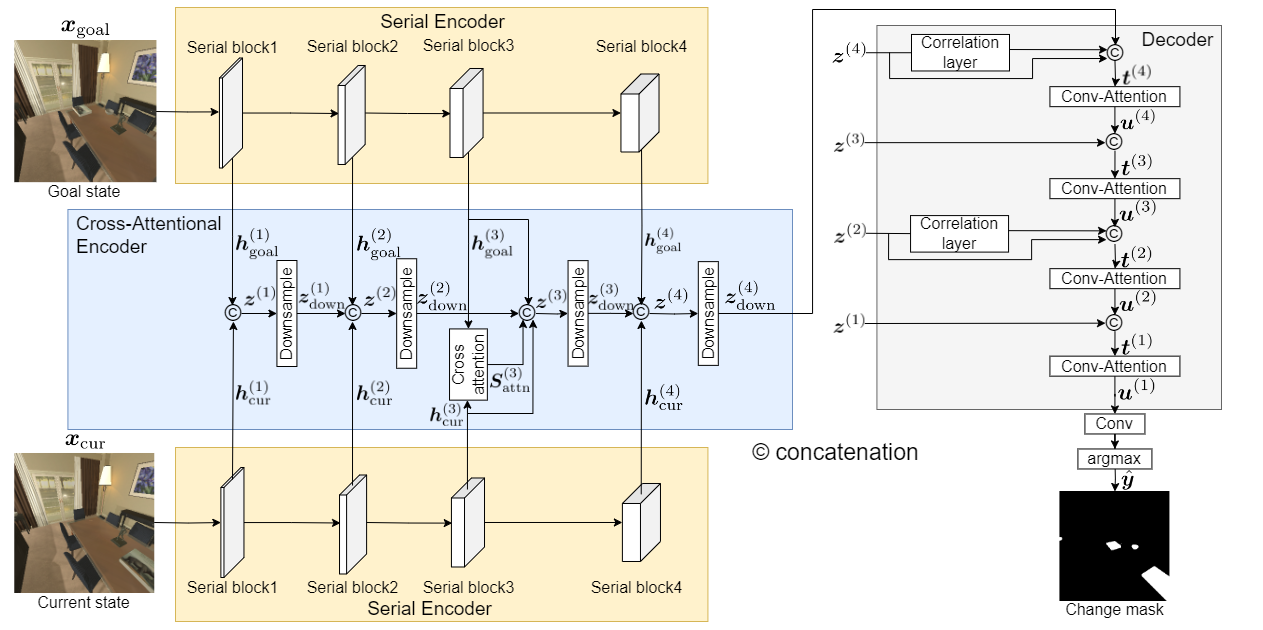}
        \caption{\normalsize The framework of our method. Our model consists of three main modules: Serial Encoder, Cross-Attentional Encoder, and Decoder.}
        \label{fig:network}
\end{figure*}

\subsection{Novelty}


In this paper, we propose the Co-Scale Cross-Attentional Transformer for RTD.
Fig.~\ref{fig:network} shows the framework of our method.
Our key contributions are the following:
\begin{itemize}
    \item We introduce the Serial Encoder that extracts features based on CoaT \cite{xu2021co} for two lanes in parallel, the goal and current states, instead of one lane.
    \item We introduce the Cross-Attentional Encoder that models the relationship between the goal and current states.
\end{itemize}

Note that our method uses attention mechanisms but is not Transformer-based.
By introducing the Serial Encoder and Cross-Attentional Encoder, we can solve problems unique to the RTD task, such as detecting changes in the angle of doors or drawers.
Our method can be applied to real-world images with the same input as in a simulation environment.



\subsection{Input}


The network inputs are defined as follows:
\begin{align*}
    \bm{x}=\{\bm{x}_{\mathrm{goal}}, \bm{x}_{\mathrm{cur}}\},
\end{align*}
where $\bm{x}_{\mathrm{goal}}\in\mathbb{R}^{3\times256\times256}$ and $\bm{x}_{\mathrm{cur}}\in\mathbb{R}^{3\times256\times256}$ denote the RGB images of the goal and current states, respectively.

\subsection{Serial Encoder}

Serial Encoder extracts image features using $M$ serial blocks \cite{xu2021co}.
The processing in the $i$-th serial block is performed as follows:
First, the input $\bm{x}_{\mathrm{goal}}$ is downsampled using a patch embedding layer to obtain $\bm{o}_{\mathrm{goal}}^{(i)}\in\mathbb{R}^{H_iW_i\times C_i}$, where $H_i$, $W_i$, and $C_i$ are the height, width, and number of channels for $\bm{o}_{\mathrm{goal}}^{(i)}$, respectively.

Next, $\bm{o}_{\mathrm{goal}}^{(i)}$ is flattened to obtain an image token.
Then, we concatenate it with the class (CLS) token \cite{xu2021co} $\bm{v}_{\mathrm{goal}}^{(i)}\in\mathbb{R}^{C_i}$ to obtain $\bm{v}_{\mathrm{token}}^{(i)}\in\mathbb{R}^{(H_iW_i+1)\times C_i}$.
Depthwise Convolution and Factorized Attention \cite{xu2021co} are applied to $\bm{v}_{\mathrm{token}}^{(i)}$ using the Conv-Attention Module.

In the serial block, we use Factorized Attention and Conv-Attention that were originally proposed in \cite{xu2021co}.
The Factorized Attention operation for an arbitrary token $\bm{X}$ is defined as follows:
\begin{align*}
    \mathrm{FactorAttn}(\bm{X}) = \frac{W_{q}\bm{X}}{\sqrt{d}} (\mathrm{softmax}(W_{k}\bm{X})^{\top})(W_{v}\bm{X}),
\end{align*}
where $W_{q}$, $W_{k}$, and $W_{v}$ are the learnable weights and $d$ is a scaling factor.
The Conv-Attention operation is defined as follows:
\begin{align*}
    \bm{\alpha} &= W_{q}\bm{X} \circ \mathrm{Depthwise}(\bm{P}, W_{v}\bm{X}), \\
    \mathrm{ConvAttn}(\bm{X}) &= \mathrm{FactorAttn}(\bm{X})+\bm{\alpha},
\end{align*}
where $\circ$ is the Hadamard product.
$\bm{P}$ and $\mathrm{Depthwise}(\cdot)$ denote positional encoding and Depthwise Convolution, respectively.
In our method, we obtain $\bm{h}_{\mathrm{goal}}^{(i)}\in\mathbb{R}^{C_i\times H_i\times W_i}$ using $\mathrm{FactorAttn}(\bm{v}_{\mathrm{token}}^{(i)})$ and $\mathrm{ConvAttn}(\bm{v}_{\mathrm{token}}^{(i)})$.

After that, the image token and CLS token are separated, and the image token is reshaped into $\bm{h}_{\mathrm{goal}}^{(i)}$.
Finally, we obtain $\{\bm{h}_{\mathrm{goal}}^{(i)} \mid i=1, \ldots , M\}$ as the outputs of this module.
Given $\bm{x}_{\mathrm{cur}}$, we obtain $\{\bm{h}_{\mathrm{cur}}^{(i)} \mid i=1, \ldots , M\}$ in a similar manner.

\subsection{Cross-Attentional Encoder}

This module extracts latent features between the goal and current states by downsampling and applying cross-attention layer by layer.
It uses $\{\bm{h}_{\mathrm{goal}}^{(i)}, \bm{h}_{\mathrm{cur}}^{(i)} \mid i=1, \ldots , M\}$ as the input.

First, $\bm{h}_{\mathrm{goal}}^{(1)}$ and $\bm{h}_{\mathrm{cur}}^{(1)}$ are concatenated to obtain  $\bm{z}^{(1)}\in\mathbb{R}^{(C_1\times 2)\times H_1\times W_1}$, as follows:
\begin{align*}
    \bm{z}^{(1)}=\{\bm{h}_{\mathrm{goal}}^{(1)}, \bm{h}_{\mathrm{cur}}^{(1)}\}.
\end{align*}

Next, $\bm{z}^{(1)}$ is downsampled to obtain $\bm{z}_{\mathrm{down}}^{(1)}\in\mathbb{R}^{C_1\times H_2\times W_2}$.
After that, $\bm{z}_{\mathrm{down}}^{(1)}$, $\bm{h}_{\mathrm{goal}}^{(2)}$, and $\bm{h}_{\mathrm{cur}}^{(2)}$ are concatenated to obtain $\bm{z}^{(2)}\in\mathbb{R}^{(C_1+C_2\times 2)\times H_2\times W_2}$, as follows:
\begin{align*}
    \bm{z}^{(2)}=\{\bm{z}_{\mathrm{down}}^{(1)}, \bm{h}_{\mathrm{goal}}^{(2)}, \bm{h}_{\mathrm{cur}}^{(2)}\}.
\end{align*}
Then, $\bm{z}_{\mathrm{down}}^{(2)}\in\mathbb{R}^{C_2\times H_3\times W_3}$ is generated by downsampling $\bm{z}^{(2)}$.

We define cross-attention using arbitrary matrices $\bm{X}_{\mathrm{A}}$ and $\bm{X}_{\mathrm{B}}$, as follows:
\begin{align*}
    \small
    \begin{split}
        \bm{f}_\mathrm{attn}^{(j)}(\bm{X}_{\mathrm{A}}, \bm{X}_{\mathrm{B}}) &=\\ \mathrm{softmax}&(\frac{(W_{q}^{(j)}\bm{X}_{\mathrm{A}})(W_{k}^{(j)}\bm{X}_{\mathrm{B}})^{\top}}{\sqrt{d_{\mathrm{CA}}}})(W_{v}^{(j)}\bm{X}_{\mathrm{B}}),
    \end{split}
\end{align*}
where $W_{q}$, $W_{k}$, and $W_{v}$ are the learnable weights.
$d_{\mathrm{CA}}$ is a scaling factor.

For $i = 3, \ldots , M-1$, we apply cross-attention to $\bm{h}_{\mathrm{goal}}^{(i)}$ and $\bm{h}_{\mathrm{cur}}^{(i)}$. 
We apply multi-head attention to them to calculate the attention score $\bm{S}_{\mathrm{attn}}^{(i)}$, as follows:
\begin{align*}
    \begin{split}
        \bm{S}_\mathrm{attn}^{(i)} =
        \{\bm{f}_\mathrm{attn}^{(j)}(\bm{h}_{\mathrm{goal}}^{(i, j)}, \bm{h}_{\mathrm{cur}}^{(i, j)}) \mid j=1, \ldots,A \},
    \end{split}
\end{align*}
where $A$ denotes the number of attention heads.

Based on the above, $\bm{z}^{(i)}\in\mathbb{R}^{(C_{i-1}+C_i\times 3)\times H_i\times W_i}$ is obtained by concatenating $\bm{z}_{\mathrm{down}}^{(i-1)}$, $\bm{S}_\mathrm{attn}^{(i)}$, $\bm{h}_{\mathrm{goal}}^{(i)}$, and $\bm{h}_{\mathrm{cur}}^{(i)}$, as follows:
\begin{align*}
    \bm{z}^{(i)}=\{\bm{z}_{\mathrm{down}}^{(i-1)}, \bm{S}_\mathrm{attn}^{(i)}, \bm{h}_{\mathrm{goal}}^{(i)}, \bm{h}_{\mathrm{cur}}^{(i)}\}.
\end{align*}
After that, $\bm{z}^{(i)}$ is downsampled to generate $\bm{z}_{\mathrm{down}}^{(i)}\in\mathbb{R}^{C_i\times H_{i+1}\times W_{i+1}}$.

Then, $\bm{z}^{(M)}\in\mathbb{R}^{(C_{M-1}+C_M\times 3)\times H_M\times W_M}$ is generated as follows:
\begin{align*}
    \bm{z}^{(M)} &=
    \{\bm{z}_{\mathrm{down}}^{(M-1)}, \bm{h}_{\mathrm{goal}}^{(M)}, \bm{h}_{\mathrm{cur}}^{(M)}\}.
\end{align*}
We apply a convolution layer to $\bm{z}^{(M)}$ and generate $\bm{z}_{\mathrm{down}}^{(M)}\in\mathbb{R}^{C_M\times H_M\times W_M}$.
Thus, this module outputs $\{\bm{z}^{(i)} \mid i=1, \ldots, M\}$ and $\bm{z}_{\mathrm{down}}^{(M)}$.

\subsection{Decoder}

In the Decoder, $M$ Conv-Attention Modules are used to obtain a predicted mask image given $\{\bm{z}^{(i)} \mid i=1, \ldots, M\}$ and $\bm{z}_{\mathrm{down}}^{(M)}$.
First, the patch embedding layer is applied to $\bm{z}_{\mathrm{down}}^{(M)}, \bm{z}^{(M)}$, and the features resulting from applying the correlation layer \cite{dosovitskiy2015flownet} to $\bm{z}^{(M)}$ in order to generate $\bm{t}^{(M)}\in\mathbb{R}^{H_M W_M\times C_{M-1}}$.
Next, the Conv-Attention Module and upsampling layer are applied to $\bm{t}^{(M)}$ to generate $\bm{u}^{(M)}\in\mathbb{R}^{C_{M-1}\times H_{M-1}\times W_{M-1}}$.

To obtain $\bm{t}^{(i)}\in\mathbb{R}^{H_i W_i\times C_{i-1}} \hspace{1mm} (i\in\{{M-1}, \ldots,\ 2\})$, we apply the patch embedding layer to $\{\bm{u}^{(i+1)}, \bm{z}^{(i)}\}$, where for $i = 2$, we use the correlation layer.
After that, the Conv-Attention Module and upsampling layer are applied to $\bm{t}^{(i)}$ to generate $\bm{u}^{(i)}\in\mathbb{R}^{C_{i-1}\times H_{i-1}\times W_{i-1}}$.
Finally, $\bm{u}^{(1)}\in\mathbb{R}^{({C_1}/2)\times 256\times 256}$ is generated from $\{\bm{u}^{(2)}, \bm{z}^{(1)}\}$.
After we apply a convolutional layer of kernel size 1 to $\bm{u}^{(1)}$, we binarize it to obtain a change mask $\hat{\bm{y}}$ of size $256\times256$, which is the final output of the model.

\subsection{Soft Dice Loss}

The loss function $\mathcal{L}$ is defined as follows:
\begin{align*}
    \mathcal{L} =     \lambda_{\mathrm{ce}}\mathcal{L}_{\mathrm{ce}}(\bm{y}, \hat{\bm{y}}) + \lambda_{\mathrm{sDice}}\mathcal{L}_{\mathrm{sDice}}(\bm{y}, p(\hat{\bm{y}})),
\end{align*}
where $\lambda_{\mathrm{ce}}$ and $\lambda_{\mathrm{sDice}}$ are loss weights.
$\mathcal{L}_{\mathrm{ce}}$ and $\mathcal{L}_{\mathrm{sDice}}$ denote the cross-entropy loss and the novel soft Dice loss, respectively.

The Soft Dice loss is newly introduced to refine under-segmented regions.
The loss is inspired by the IoU loss \cite{rahman2016iou} and Dice loss \cite{milletari2016dice}.
Unlike Dice loss \cite{milletari2016dice}, the soft Dice loss incorporates a soft assignment of mask predictions instead of a hard assignment.
$\mathcal{L}_{\mathrm{sDice}}$ is defined as follows:
\begin{align*}
    \mathcal{L}_{\mathrm{sDice}} = 1 - \frac{1}{K} \sum_{i=1}^K\frac{2\sum_{j=1}^N{\bm{y}}_{i, j}p(\hat{\bm{y}}_{i, j})}{\sum_{j=1}^N{\bm{y}}_{i, j} + \sum_{j=1}^N p(\hat{\bm{y}}_{i, j}) + \epsilon},
\end{align*}
where $K$, $N$, $\bm{y}_{i, j}$, and $\hat{\bm{y}}_{i, j}$ are the number of classes, the number of total pixels, the $j$-th pixel value of the $i$-th class in the ground-truth change mask, and that in the predicted change mask, respectively. 
$\epsilon$ is a small positive value used to avoid zero division and is set to $1\times 10^{-7}$.
\begin{table}[t]
    \centering
    \caption{\normalsize Experimental setup}\label{tab:param}
    \vspace{2mm}
    {\normalsize
    \begin{tabular}{l|l}
    \hline
    Optimizer & Adam ($\beta_1=0.5$, $\beta_2=0.999$) \\
    \hline
    Learning rete & $0.001$ \\
    \hline
    Batch size & $16$ \\
    \hline
    Loss weights & $\lambda_{\mathrm{ce}}=1, \lambda_{\mathrm{sDice}}=1$ \\
    \hline
    Serial Encoder & \#L:[2, 2, 2, 2], \#A:8 \\
    \hline
    Cross-Attentional & \multirow{2}{*}{\#A:8} \\
    Encoder &  \\
    \hline
    Decoder & \#L:[1, 1, 1, 1], \#A:8 \\
    \hline
    \end{tabular}
    }
\end{table}

\section{Experimental Setup
\label{dataset}
}


In this paper, we built a new RTD dataset in a simulation environment.
This is because existing datasets (e.g., PCD \cite{sakurada2015pcd} and ChangeSim \cite{park2021changesim}) were built neither for rearrangement tasks nor in the domestic environment and are not suitable for our target task.
Instead, we used AI2-THOR \cite{ai2thor} to build this dataset using the following procedure.

First, we collected egocentric images corresponding to the robot’s current viewpoint in a room.
Second, we randomly moved objects and changed door angles.
In practical scenarios, the robot pose in the current state may not be identical to that in the goal state.
To account for this, we randomly added noise to the orientation of the robot by $+\delta$ or $-\delta$ degrees ($\delta = 2$).
Third, we again collected the images seen by the robot in the room.
When collecting the RGB images, we saved the IDs and positions of the rearranged objects for each of the goal and current states.
Using these, we obtained images where the pixels of the rearrangement target objects were masked in the current state.
We define rearrangement target objects as objects that have been moved more than 30 centimeters or doors/drawers whose angle has been changed by more than 60\% of their maximum opening angle.

\begin{table*}[!t]
    \caption{\normalsize Quantitative results and ablation studies.}
    \vspace{2mm}
    \centering
    {\normalsize
    \begin{tabular}{c|c|c|c|c|c} \hline
        \multicolumn{2}{c|}{\multirow{2}{*}{Model}} & Serial Encoder & \multirow{2}{*}{Attention} & \multirow{2}{*}{$F_1$-score [\%]} & \multirow{2}{*}{mIoU [\%]} \\
        \multicolumn{2}{c|}{} & Type & {} & & \\ \hline\hline
        \multicolumn{2}{c|}{Pixel difference} & - & - & $34.1$ & $1.8$ \\ \hline
        \multicolumn{2}{c|}{CSCDNet \cite{sakurada2020weakly}} & - & - & $76.8\ {\tiny\pm}\ 2.5$ & $42.0\ {\tiny\pm}\ 4.4$ \\ \hline
        \multirow{4}{*}{Ours} & (i) & Res block \cite{he2016deep} & CA & $72.9\ {\tiny\pm}\ 0.5$ & $35.2\ {\tiny\pm}\ 0.8$ \\ \cline{2-6}
         & (ii) & Serial Block \cite{xu2021co} & - & $76.4\ {\tiny\pm}\ 0.3$ & $41.7\ {\tiny\pm}\ 0.7$ \\ \cline{2-6}
         & (iii) & Serial Block \cite{xu2021co} & CA+SA & $77.9\ {\tiny\pm}\ 2.3$ & $44.2\ {\tiny\pm}\ 4.3$ \\ \cline{2-6}
         & (iv) & Serial Block \cite{xu2021co} & CA & $\bm{80.3\ {\tiny\pm}\ 0.2}$ & $\bm{48.6\ {\tiny\pm}\ 0.5}$ \\ \hline
    \end{tabular}
    \label{tab:quan&ablation}
    }
\end{table*}

RTD dataset includes 12,000 samples. Each sample consists of RGB images of the goal and current states, and a change mask.
In the RTD dataset, the training, validation, and test sets consist of 10,000, 1,000, and 1,000 samples, respectively.
Notably, the environments for the training, validation, and test set were not overlapped.
In addition, the test set was collected in unseen environments.
We used the training set to update our model’s parameters and the validation set to tune the hyperparameters. We evaluated our model on the test set.


The experimental setup is summarized in Table~\ref{tab:param}.
\#L represents the number of layers for the Conv-Attention Module in the Serial Encoder’s serial block or Decoder, and \#A represents the number of attention heads in each module.
Note that the $i$-th value from the left of \#L indicates the number of layers of the $i$-th serial block or Conv-Attention Module.
$H_i$, $W_i$, and $C_i$ denote the height, width, and number of channels of the image feature map output from the $i$-th serial block of the Serial Encoder, respectively.
Here, we set $H_1 = W_1 = C_1 = 64$ and $(H_{i+1}, W_{i+1}, C_{i+1})=({H_i}/2, {W_i}/2, {C_i}\times2)$.
Note that $C_3 = 320$ and $C_4 = 512$.

The total number of trainable parameters was 25M, and the number of the total multiply-add operations was 990M.
We trained our model on a GeForce RTX 3090 with 24GB of memory and an Intel Core i9-10900KF with 64GB of memory.
It took 4 hours to train our model. The inference time was approximately 18 milliseconds for one sample.
We evaluated the validation loss and stopped training if the loss did not improve for five consecutive epochs.
The final performance of the test set was obtained when the validation loss was minimized.

As the evaluation metrics, we used $F_1$-score and mIoU.
The primary metric is the mIoU.
The mIoU and $F_1$-score are defined as follows:
\begin{align*}
    \mathrm{mIoU} &= \frac{1}{N}\sum_{i=1}^{N}\mathrm{IoU}(\hat{\bm{y}}_i, \bm{y}_i), \\
    F_1 &= \frac{2}{1/\mathrm{precision} + 1/\mathrm{recall}}, \\
    \mathrm{precision} &= \frac{\mathrm{TP}}{\mathrm{TP + FP}}, \\
    \mathrm{recall} &= \frac{\mathrm{TP}}{\mathrm{TP + FN}},
\end{align*}
where $N$ denotes the number of samples. $\hat{\bm{y}}_i$ and $\bm{y}_i$ denote the predicted and ground-truth masks in the $i$-th sample, respectively.
IoU denotes the Intersection over Union (IoU) between two masks.
TP, FP, and FN denote the number of true positives, false positives and false negatives, respectively.
We selected the $F_1$-score and mIoU because they are standard metrics for SCD, a closely related task to RTD.
\section{Experimental Results}

\subsection{Quantitative Results}


Table~\ref{tab:quan&ablation} shows the quantitative results.
We conducted the experiment five times.
The averages and standard deviations of the $F_1$-score and mIoU are shown in the table.
The ``pixel difference'' method represents a simple method based on the comparison of the pixel values.
We selected CSCDNet as another baseline method because it has been successfully applied to SCD tasks.
CA and SA represent cross-attention and self-attention, respectively.

Table~\ref{tab:quan&ablation} shows that the proposed method achieved the $F_1$-score of 80.3, whereas the CSCDNet \cite{sakurada2020weakly} and pixel difference method achieved 76.8 and 34.1, respectively. 
Moreover, the proposed method achieved the mIoU of 48.6, whereas the CSCDNet and pixel difference method achieved 42.0 and 1.8, respectively.
Therefore, our method outperformed the CSCDNet by 3.5 and 6.6 on the $F_1$-score and mIoU, respectively.


\begin{figure*}[t]
    \vspace{2mm}
    \centering
    \includegraphics[width=\linewidth]{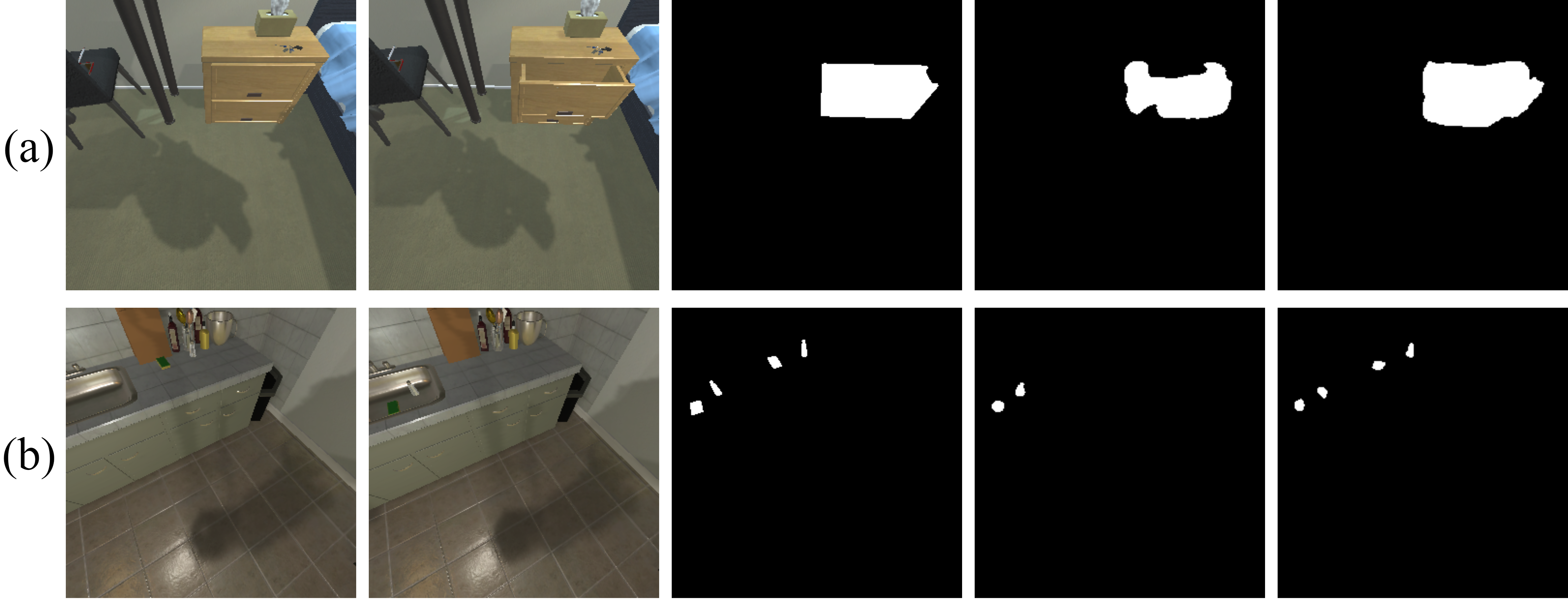}
    \vspace{-8.5mm}
    \caption{\normalsize Qualitative results of successful samples. From left to right: $\bm{x}_{\mathrm{goal}}$, $\bm{x}_{\mathrm{cur}}$, $\bm{y}$, $\hat{\bm{y}}$ obtained by the CSCDNet, and $\hat{\bm{y}}$ obtained by the proposed method. (a) The proposed method generated an almost complete segmentation of the drawer shape. (b) The proposed method successfully generated masks of all small objects.}
    \label{fig:quali}
\end{figure*}
\begin{figure*}[t]
    \centering
    \small \ \ \ 
        \includegraphics[width=0.96\linewidth]{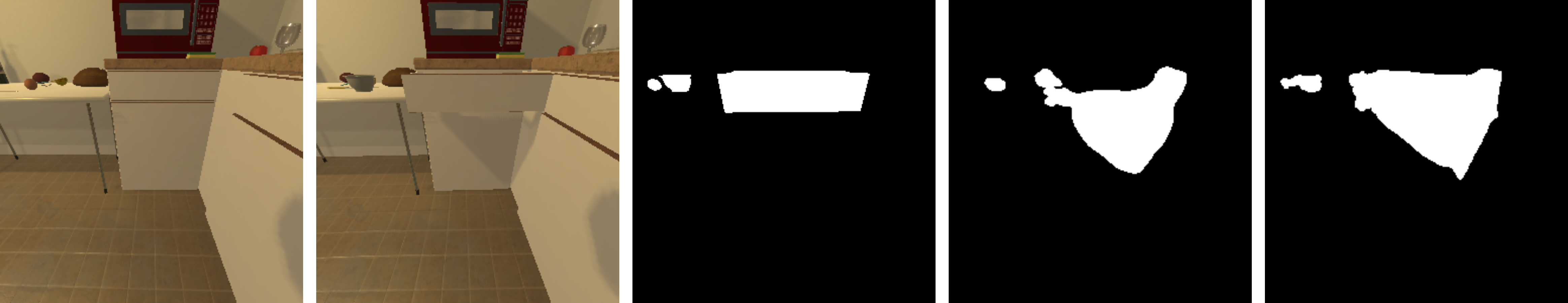}
    \vspace{-3.8mm}
    \caption{\normalsize Qualitative results of a failed sample. From left to right: $\bm{x}_{\mathrm{goal}}$, $\bm{x}_{\mathrm{cur}}$, $\bm{y}$, $\hat{\bm{y}}$ obtained by the CSCDNet, and $\hat{\bm{y}}$ obtained by the proposed method. Both the CSCDNet and the proposed method incorrectly segmented the shadows caused by the opening of the drawers.}
    \label{fig:failed}
\end{figure*}

\subsection{Qualitative Results}

Fig.~\ref{fig:quali} shows the qualitative results of successful samples. 
In the figure, the first, second, third, fourth, and fifth columns show $\bm{x}_{\mathrm{goal}}$, $\bm{x}_{\mathrm{cur}}$, $\bm{y}$, $\hat{\bm{y}}$ obtained by the CSCDNet, and $\hat{\bm{y}}$ obtained by the proposed method, respectively.

Fig.~\ref{fig:quali}(a) shows a successful sample where the degree of opening drawer was changed.
The CSCDNet failed to mask the whole area of the drawer.
In contrast, the proposed method almost completely detected the shape of the rearrangement target object.
Fig.~\ref{fig:quali}(b) also shows a successful sample where the sponge and sugar bottle were moved.
The CSCDNet failed to detect the rearrangement target objects in the goal state.
In contrast, the proposed method successfully generated masks of the rearrangement target objects in both states.
Processing each of the input images with three Serial Blocks, extracting essential features from each, and then modeling the relationship between these features through Cross-Attention has contributed to the successful detection of changes in the state of doors, drawers, and other objects.

Fig.~\ref{fig:failed} shows the qualitative results of a failed sample where the degree of opening drawer was changed.
In the current state, the drawer has shadows under it and pixel values changed.
Therefore, both the CSCDNet and the proposed methods incorrectly segmented the shadows.

\subsection{Ablation Studies}

We set the following three ablation conditions:
\begin{enumerate}
    \item[(i)] Serial block ablation: In the Serial Encoder, we used Res blocks \cite{he2016deep} instead of Serial Blocks \cite{xu2021co} serial blocks to investigate their effects on performance.
    Table~\ref{tab:quan&ablation} shows that the $F_1$-score and mIoU in Model (i) were 72.9\% and 35.2\%, respectively, while the $F_1$-score and mIoU in our method were 80.3\% and 48.6\%, respectively.
    This indicates that feature extractor using CoaT serial blocks contributed to the performance improvement.
    \item[(ii)] Cross-attention ablation: We removed the cross-attention in Cross-Attentional Encoder (see Fig.~\ref{fig:network}) to investigate whether the application of the cross-attention was effective.
    Table~\ref{tab:quan&ablation} also shows that the $F_1$-score were 80.3\% and 76.4\%, respectively, and the mIoU for our method and Model (ii) were 48.6\% and 41.7\%, respectively.
    This indicates that the removal of the cross-attention resulted in lower performance.
    This means that the application of the cross-attention was effective in modeling the relationship between the goal and current states.
    \item[(iii)] Direct-concatenation ablation: We added self-attention after the cross-attention to investigate the effectiveness of using the attention score, which was directly used for concatenation with other features.
    Table~\ref{tab:quan&ablation} shows that the $F_1$-score and mIoU in Model (iii) were 77.9\% and 44.2\%, respectively, while the $F_1$-score and mIoU in Model (iv) were 80.3\% and 48.6\%, respectively.
    This indicates that the direct use of the attention score was effective.
\end{enumerate}

Based on the results of the ablation studies, the newly introduced Serial Encoder and  Cross-Attentional Encoder can extract effective features from the two input images and model the relationship between the goal and the current state.
Furthermore, the combination of these encoders enables the detection of changes in the angle of doors or drawers, as demonstrated in Section 6.2.
Overall, the proposed method outperformed CSCDNet, as shown in Section 6.1.



\begin{table}[t]
    \begin{center}
    \caption{\normalsize Error analysis on failure cases.}
    \vspace{2mm}
    \label{tab:error_analysis}
    {\normalsize
    \begin{tabular}{m{6cm}|Wc{2cm}} \hline
    Errors                          & \# of Errors \\ \hline \hline
    Under- or over-segmentation      & 58 \\ \hline
    Too small object                 & 44 \\ \hline
    Incomplete object segmentation   & 18 \\ \hline 
    Annotation error                 & 7 \\ \hline
    Below threshold                  & 9 \\ \hline
    Light/shadow detection error     & 5 \\ \hline
    \end{tabular}
    }
  \end{center}
\end{table}

\subsection{Error Analysis and Discussion}
On the test set, there were 100 samples for which the IoU was less than 10\%.
We analyzed the errors for the 100 samples, in the order of smallest IoU to largest, in the test set.
Table~\ref{tab:error_analysis} shows the categorization of the failure cases.
If there were multiple errors for a sample, we treated each as an independent failure case.
The causes of failure can be roughly divided into six types:
\begin{itemize}
    \item Under- or over-segmentation: The generated change mask failed to cover the whole ground-truth area of the rearrangement target object sometimes because the rearrangement target object was similar in color to its background.
    \item Too small object: This category includes the cases where our method could not capture the change regions of an object whose size was less than 1\% of the image area.
    \item Incomplete object segmentation: This category includes the cases where our method could not capture change regions of the rearrangement target objects that were not completely in the image.
    \item Annotation error: The ground-truth change mask was inappropriate. For example, a rearrangement target object was not masked because it was outside the image in either the goal or current states.
    \item Below threshold: The generated change mask included an object that moved less than 30 centimeters and was not treated as a rearrangement target object.
    \item Light/shadow detection error: The generated change mask incorrectly included the light or shadow region of the rearrangement target object.
\end{itemize}

Table~\ref{tab:error_analysis} indicates that the main bottleneck was ``Under- or over-segmentation.''
In order for the model to correctly capture the change regions of rearrangement target objects, using SAM \cite{kirillov2023segment} as a supplementary preprocessing step may be a solution for understanding the size of objects in the image.
\vspace{-2mm}
\section{Conclusions}

In this paper, we focused on RTD, in which the model detects rearrangement target objects that were moved between the goal and current states in the room where the robot is located.

We would like to emphasize the following contributions of this study:
\begin{itemize}
    \item We proposed the Co-Scale Cross-Attentional Transformer for RTD. 
    \item We introduced the Serial Encoder that extracts features based on CoaT \cite{xu2021co} for two lanes in parallel, the goal and current states, instead of one lane.
    \item We introduced the Cross-Attentional Encoder that models the relationship between the goal and current states.
    \item For the RTD dataset, our method outperformed the baseline method for all the metrics.
\end{itemize}

In future work, we plan to apply our model to real-world images obtained by physical robots that conduct rearrangement tasks.
We also plan to extend our model and develop new methods applicable to SCD tasks.

\section*{ACKNOWLEDGMENT}
\vspace{-1mm}
This work was partially supported by JSPS KAKENHI Grant Number 20H04269, JST Moonshot, and NEDO.
\vspace{-1mm}
\bibliographystyle{tfnlm}
\bibliography{reference}
\end{document}